\documentclass{article}

\usepackage[main, final]{neurips/neurips_2026}

\usepackage{amsmath, amssymb, amsthm}
\usepackage{graphicx}
\graphicspath{{figures/}}
\usepackage{booktabs}
\usepackage{hyperref}
\usepackage{multirow}
\usepackage{array}
\usepackage{subcaption}
\usepackage{float}
\usepackage{xcolor}
\usepackage{bm}
\usepackage{tikz}
\usetikzlibrary{arrows.meta, positioning, fit, backgrounds, calc}
\usepackage{titlesec}

\titlespacing*{\section}{0pt}{8pt plus 2pt minus 2pt}{4pt plus 1pt minus 1pt}
\titlespacing*{\subsection}{0pt}{6pt plus 1pt minus 1pt}{3pt plus 1pt minus 1pt}
\titlespacing*{\paragraph}{0pt}{4pt plus 1pt minus 1pt}{1em}

\DeclareMathOperator{\Var}{Var}

\DeclareMathOperator{\E}{E}

\title{Hidden Measurement Error in LLM Pipelines Distorts Annotation, Evaluation, and Benchmarking\thanks{The SI Appendix is available at \url{https://solomonmg.github.io/pdf/totalevalerror_si.pdf}.}}

\author{%
  Solomon Messing\thanks{Corresponding author.} \\
  Center for Social Media, AI, and Politics, New York University \\
  MLCommons \\
}


\begin{document}

\maketitle

\begin{abstract}
LLM evaluations drive model deployment, safety standards, research conclusions, and future projections, for example, about AI's impact on the labor market. Yet the way we often estimate noise from autoraters (LLM-as-judge) ignores variability from factors like judge model choice and prompt phrasing, which means we're underestimating error in a way that \emph{worsens} with more data. This unmeasured variance can affect results enough to reverse conclusions \citep{baumann2025llmhacking, huang2026dropping}. And judge-pipelines that fail to average this variation are more vulnerable to ``benchmark hacking'' \citep{singh2025leaderboard}. This paper introduces a framework we call ``total evaluation error (TEE),'' which decomposes LLM pipeline uncertainty and uses design-study projections to reduce total error. In our safety and ideology applications, naive standard errors are 53 - 70\% smaller than the corrected SE under our TEE framework. Using Chatbot Arena data, we show naive 95\% CI coverage drops as $n$ grows, while TEE-corrected coverage holds at 95\%, and TEE-guided pipelines restrict the potential range of benchmark gaming by roughly 43\%. We show further that a small pilot allows for the estimation of proper CIs, with the added benefit of projecting which design changes most improve precision. Using those projections to allocate API calls cut MMLU estimation error roughly in half against the answer key at the same budget, and increased per-match agreement with human voters on Chatbot Arena by 5.5 percentage points.
\end{abstract}

\section{Introduction}
\label{sec:intro}

LLM measurement pipelines introduce variance at multiple stages that traditional reporting norms ignore. Researchers who use LLMs for annotation, evaluation, or content moderation routinely do so with no assessment of prompt sensitivity, temperature dependence, or item-level heterogeneity \citep{pecher2026underspecification, barrie2025replication}. Benchmark scores likewise depend on prompt wording and scoring model choice, and this dependence is seldom measured (SI Appendix, Table~\ref{tab:current_practice}).\footnote{AlpacaEval and Chatbot Arena report item-sampling CIs; AILuminate v1.0 used a multi-judge evaluator ensemble (since v1.2, a deterministic evaluator trained by MLCommons); no benchmark decomposes prompt, temperature, or interaction variance. Full survey of 10 benchmarks and 3 annotation studies in SI Appendix, Table~\ref{tab:current_practice}.} In all cases, unmeasured variation can create analogues to $p$-hacking \citep{simmons2011falsepositive, gelman2013garden} and overfitting \citep{recht2019imagenet} such that the results reflect measurement noise, not general phenomena.

\begin{figure}[htbp]
\centering
\includegraphics[width=0.9\linewidth]{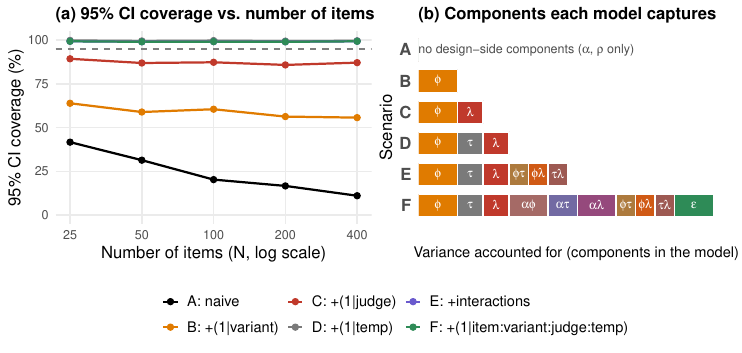}
\caption{Naive confidence intervals fail because pipeline variance from prompt, judge, and temperature choice does not shrink with items ($N$). (a)~95\% CI coverage versus $N$ in 6 scenarios (A-F). (b)~Variance components accounted for in each scenario. Scenario A corresponds to the naive estimate $\Phi^{-1}(0.975) \times s/\sqrt{N}$; scenarios B-F represent adding more variance parameters estimated according to the framework introduced below. Scenarios D-F overlap.%
}
\label{fig:underestimation}
\end{figure}

At best, current practice reports confidence intervals (CIs) reflecting variation from repeated sampling at identical settings, ignoring prompt sensitivity, item heterogeneity, and their interactions. Figure \ref{fig:underestimation} presents simulation results showing that we should expect under-coverage to \emph{worsen} with more items, because omitted variance from prompt and judge choices does not shrink with~$N$.\footnote{Variance parameters estimated via REML according to the framework introduced below; 1,000 simulation replicates per data point (see Section~\ref{si:simulations} for simulation details).}

The problem is especially urgent in benchmarking, where unstable scores make rank-order differences between closely spaced models (or systems under test, SUT) uninterpretable for evaluators and give model developers exploitable degrees of freedom. On stable measurement, \citet{romanou2026brittlebench} show that semantics-preserving input perturbations explain near half of performance variance, minor reorderings of answer choices shift MMLU leaderboard rankings by up to 8~positions \citep{alzahrani2024benchmarks}, and on Chatbot Arena dropping just 0.003\% of human preferences is enough to flip the top-ranked model \citep{huang2026dropping}. With respect to exploitation, \citet{singh2025leaderboard} document best-of-$K$ leaderboard gaming in production, the mechanism Section~\ref{sec:gaming} quantifies.

Ignoring this variability can harm the scientific literature as well. One re-evaluation study found 31\% of hypotheses tested with state-of-the-art LLM annotations yielded incorrect conclusions under LLM configuration variation (model, prompt), and found malicious actors could reverse the sign of 68\% of statistically significant effects through adversarial configuration selection \citep{baumann2025llmhacking}. Other work has found that formatting choices alone produced a 76-point accuracy spread \citep{sclar2024sensitivity}, while infrastructure nondeterminism produced accuracy standard deviations of up to nine percentage points even at temperature zero \citep{yuan2025numerical}.\footnote{Note that setting $T = 0$ does not eliminate underlying model uncertainty. Greedy decoding picks the highest-probability token at each step to yield a single deterministic output, even when the token distribution has high entropy, hiding the variation that sampling at $T > 0$ would show. In chain-of-thought models, greedy decoding across a long reasoning trace can also lock in early token choices that do not reflect the broader distribution of answers \citep{renze2024temperature}.}

\section{Related Work}
\label{sec:related}

Several concurrent works address components of this problem, including variance decomposition \citep{haase2026withinmodel, wang2024measuring}, uncertainty quantification \citep{longjohn2025statistical}, sensitivity taxonomies \citep{romanou2026brittlebench}, meta-evaluation of benchmark reliability \citep{mcgregor2025benchrisk}, and statistical tests for whether LLMs can replace human annotators \citep{calderon2025alttest}. \citet{camuffo2026variance} develop a variance taxonomy for LLM annotation (construct, interface, model, output, aggregation) and tie annotation variance to bias in downstream estimates. NIST AI 800-3 \citep{keller2026nist} develop a GLMM for benchmark evaluation, distinguish benchmark- from generalized accuracy as separate estimands, decompose variance into inter- and intra-item components, and propose a number of design diagnostics. However, \citet{dobriban2025statistical} argue that ``a comprehensive and unified statistical methodology that addresses most of the common evaluation problems \ldots remains to be developed.'' TEE builds on generalizability theory (G-theory) \citep{cronbach1972dependability, brennan2001gtheory, shavelson1991gtheory, bayerl2007gtheory, song2025gtheory} and uses its standard estimators. We contribute the LLM-specific measurement design that makes prompts, temperatures, models, judges, scoring methods, and interactions such as item$\times$judge and prompt$\times$judge explicit sources of pipeline error, where they are currently hidden implementation choices (SI Appendix, Section~\ref{si:gtheory} maps G-theory terms onto TEE).

\section{Total Evaluation Error (TEE) Framework}
\label{sec:framework}

We call this framework \textit{Total Evaluation Error} (TEE, Figure~\ref{fig:pipeline}). It decomposes the variance using first order interactions in a mixed effects model. The framework also yields projections that can be used to generate better calibrated CIs and reduce error through aggregation using a D-study. The name follows total survey error \citep{sen2019tedon, groves2010survey}, which treats every stage of a measurement pipeline as a source of error, and the framework draws on the text-as-data literature \citep{grimmer2013text}. Note that TEE focuses on measurement reliability, whereas judge correctness and construct coverage are validity questions that require external ground-truth data (Sections~\ref{sec:safety} and~\ref{sec:limitations}). %

\begin{figure}[t!]
\centering
\resizebox{0.82\linewidth}{!}{%
\begin{tikzpicture}[
    node distance=0.4cm and 1.6cm,
    box/.style    = {rectangle, draw, thick, rounded corners=2pt,
                     minimum width=4.0cm, minimum height=0.7cm,
                     align=center, font=\small, fill=blue!4},
    final/.style  = {box, fill=blue!18, font=\bfseries\small,
                     minimum width=4.4cm},
    layer/.style  = {draw=gray!55, dashed, semithick, rounded corners=5pt,
                     inner xsep=10pt, inner ysep=4pt, fill=gray!5},
    llabel/.style = {font=\scriptsize\itshape, text=gray!72!black,
                     fill=white, inner xsep=2pt, inner ysep=0pt},
    vleft/.style  = {font=\footnotesize, align=right, anchor=east,
                     text width=4.6cm},
    vright/.style = {font=\footnotesize, align=left,  anchor=west,
                     text width=4.6cm},
    arrow/.style  = {-{Stealth[length=2.6mm]}, thick},
    header/.style = {font=\small\bfseries, align=center,
                     text=gray!40!black, text width=4.6cm}
]

\node[box]                            (items)  {\textbf{Items} {\scriptsize\itshape(benchmark or annotation)}};
\node[box, below=0.55cm of items]     (sutpr)  {\textbf{SUT prompt} {\scriptsize\itshape(instruction wrapping item)}};
\node[box, below=of sutpr]            (sutres) {\textbf{SUT response} {\scriptsize\itshape(model's answer)}};
\node[box, below=0.55cm of sutres]    (jpr)    {\textbf{Judge prompt} {\scriptsize\itshape(rubric, scale, examples)}};
\node[box, below=of jpr]              (jres)   {\textbf{Judge rating} {\scriptsize\itshape(judge's score)}};
\node[final, below=0.55cm of jres]    (agg)    {Aggregate score $\hat{\theta}_{hm}$};

\draw[arrow] (items)  -- (sutpr);
\draw[arrow] (sutpr)  -- (sutres);
\draw[arrow] (sutres) -- (jpr);
\draw[arrow] (jpr)    -- (jres);
\draw[arrow] (jres)   -- (agg);

\begin{scope}[on background layer]
  \node[layer, draw=gray!40, fill=gray!3,
        fit=(sutpr)(sutres)] (sutlayer) {};
  \node[layer, fit=(jpr)(jres)]      (jlayer)   {};
\end{scope}
\node[llabel, anchor=north] at ([yshift=-1pt]sutlayer.north)
   {SUT layer (skip for content annotation)};
\node[llabel, anchor=north] at ([yshift=-1pt]jlayer.north)
   {judge / annotation layer};

\node[vleft] at ([xshift=-1.0cm]items.west)
    {real differences between items {\scriptsize\textcolor{gray!55!black}{$\sigma^{2}_{\alpha}$}}};
\node[vleft] at ([xshift=-1.0cm]sutpr.west)
    {rewording the instruction changes answers {\scriptsize\textcolor{gray!55!black}{$\sigma^{2}_{\phi}$}}};
\node[vleft] at ([xshift=-1.0cm]sutres.west)
    {noise across replicate calls {\scriptsize\textcolor{gray!55!black}{$\sigma^{2}_{\rho}$}}};
\node[vleft] at ([xshift=-1.0cm]jpr.west)
    {rewording the rubric changes scores {\scriptsize\textcolor{gray!55!black}{$\sigma^{2}_{\phi}$}}};
\node[vleft] at ([xshift=-1.0cm]jres.west)
    {judges disagree on which items get scored how {\scriptsize\textcolor{gray!55!black}{$\sigma^{2}_{\alpha\lambda}$}}};

\node[vright] at ([xshift=1.0cm]items.east)
    {category sampling, sample size $N$};
\node[vright] at ([xshift=1.0cm]sutpr.east)
    {format, system message, few-shot examples};
\node[vright] at ([xshift=1.0cm]sutres.east)
    {temperature, seed, sampling parameters};
\node[vright] at ([xshift=1.0cm]jpr.east)
    {rubric structure, scale anchors};
\node[vright] at ([xshift=1.0cm]jres.east)
    {which judge model, scoring rule (binary / Likert / BT)};

\node[header] at ([xshift=-5.2cm,yshift=0.5cm]items.north)
    {sources of noise\\{\normalfont\footnotesize shrink with more data}};
\node[header] at ([xshift=5cm,yshift=0.5cm]items.north)
    {design choices\\{\normalfont\footnotesize aggregate/sensitivity analysis}};
\end{tikzpicture}}%

\caption{TEE separates aggregation-reducible noise from design-choice sensitivity at every pipeline stage. %
For content-annotation tasks (e.g., labeling social media posts) the SUT layer is omitted.}
\label{fig:pipeline}
\end{figure}

Figure~\ref{fig:safety_combined}(a) illustrates the TEE decomposition based on 141 safety items from the AILuminate benchmark. The 3-judge panel contributes 43.8\% of the variance with item$\times$judge adding another 16.7\%. On the other hand, prompt wording contributes less than 0.5\%. This means investing in additional judges will lower the variance more than prompts or additional benchmark items.

\subsection{Data-Generating Process}
\label{sec:dgp}

An LLM evaluation pipeline yields output $Y_{ivhm}^{(r)}$, based on item $i$ ($i = 1, \ldots, N$), prompt variant $v$ ($v = 1, \ldots, V$), temperature $h$, model $m$ (a judge or SUT), and replication $r$. Replications are repeated model calls.\footnote{In G-theory terms, these factors and the scoring protocol comprise the admissible observations.} For agents, the harness sets the system prompt, the tools, and the agent loop. Different harnesses therefore count as different prompt variants, and the prompt terms below include harness variance. The framework specifies the linear mixed model:

\begin{equation}
\label{eq:dgp}
\begin{split}
Y_{ivhm}^{(r)} = \mu + \alpha_i + \phi_v + \tau_h + \lambda_m 
 + (\alpha\phi)_{iv} + (\alpha\tau)_{ih} + (\phi\tau)_{vh} 
 + (\alpha\lambda)_{im} + (\phi\lambda)_{vm} \\
 + \epsilon_{ivhm} + \rho_{ivhm}^{(r)}
\end{split}
\end{equation}

where $\mu$ is the grand mean, $\alpha_i$ is the item random effect, $\phi_v$ the prompt variant random effect, $\tau_h$ the temperature fixed effect, $\lambda_m$ the model fixed effect, and the five parenthesized interaction terms are two-way interaction random effects. $\epsilon_{ivhm} \sim N(0, \sigma^2_\epsilon)$ captures cell-level error from unmodeled higher-order interactions, and $\rho_{ivhm}^{(r)} \sim N(0, \sigma^2_\rho)$ captures within-cell replicate noise. %

We treat items and prompt variants as \emph{random} to get an estimand over all items and prompts that might have been drawn. On the other hand, we treat temperature and model as \emph{fixed} because usually we select judge models with intent and generally do not seek to generalize to some model population. Similarly we consider temperatures as analogous to a ``dose'' rather than something to be sampled. Note that temperature has nonlinear, non-additive effects on output variance \citep{renze2024temperature, li2025hotcold} (SI Appendix, Section~\ref{si:temperature}). %

\subsection{Variance Decomposition}
\label{sec:gstudy}

Decomposing the variance requires three formal assumptions: conditional exchangeability of replications and prompt variants, additivity with two-way interactions, and independent zero mean random effects (SI Appendix, Section~\ref{si:framework}). %

For prompt variants, the exchangeability assumption requires that variants all measure the same construct. Drawing on G-theory \citep{cronbach1972dependability, brennan2001gtheory}, survey methodology on question wording \citep{groves2010survey}, and multi-prompt LLM evaluation \citep{mizrahi2024multiprompt, sclar2024sensitivity, barrie2024pss}, we require that variants hold constant the task or rubric, inserted benchmark content, answer format, reasoning level, and scoring mode (SI Appendix, Section~\ref{si:prompt_admissibility}).

Because the random effects are mutually independent, their variances add. At fixed temperature $h$ and model $m$, $\tau_h$ and $\lambda_m$ are constants and drop out, so the observation-level variance decomposes as:
\begin{equation}
\label{eq:tle}
\begin{split}
\Var(Y \mid h, m) = \sigma^2_\alpha + \sigma^2_\phi + \sigma^2_{\alpha\phi} + \sigma^2_{\alpha\tau} \\
+ \sigma^2_{\phi\tau} + \sigma^2_{\alpha\lambda} + \sigma^2_{\phi\lambda} + \sigma^2_\epsilon + \sigma^2_\rho
\end{split}
\end{equation}

At any fixed level, the temperature and model interaction terms still vary across the sampled items and prompts, so they remain in the decomposition.

Random components (item, prompt variant, replication) shrink by aggregation within a chosen design; fixed-effect contributions (temperature, model, scoring method) shrink only by aggregating across designs, and otherwise should be reported as sensitivity alongside the point estimate \citep{steegen2016multiverse}.

\subsection{D-Study Projections}
\label{sec:dstudy}

Once variance components are estimated, a D-study projects the variance of the estimated grand mean $\hat{\theta}$ under a proposed design. We denote the \emph{proposed} D-study design ($N'$, $V'$, $R'$), while the observed G-study counts remain $N$, $V$, $R$, which are used to estimate the variance components. %

 In classical psychometrics with human raters, D-studies involve trying to choose how many raters, items, and occasions a study can afford \citep{brennan2001gtheory}. However, in LLM/autorater pipelines how much of each facet (e.g., $N'$ items, $V'$ variants, $R'$ replicates) a study can afford is a matter of API calls, so an autorater D-study can trade off all facets under a given compute budget.

For $N'$ items, $V'$ prompt variants, and $R'$ replications at fixed temperature, the variance is:\footnote{When averaging over $M$ models, a design-sensitivity term $\sigma^2_\lambda = \frac{1}{M}\sum_m(\lambda_m - \bar{\lambda})^2$ is added. The same is true for averaging over multiple temperatures. Here $\bar{\sigma}^2_\epsilon = \frac{1}{M}\sum_m \sigma^2_{\epsilon,m}$ and $\bar{\sigma}^2_\rho = \frac{1}{M}\sum_m \sigma^2_{\rho,m}$ are the cell-level and replicate-noise variances averaged over models.}
\begin{equation}
\label{eq:dstudy_multi}
\begin{split}
\Var(\hat{\theta}_h) = \frac{\sigma^2_\alpha}{N'} + \frac{\sigma^2_\phi}{V'} + \frac{\sigma^2_\lambda}{M} + \frac{\sigma^2_{\alpha\phi}}{N'V'} + \frac{\sigma^2_{\alpha\tau}}{N'} + \frac{\sigma^2_{\phi\tau}}{V'} \\
+ \frac{\sigma^2_{\alpha\lambda}}{N'M} + \frac{\sigma^2_{\phi\lambda}}{V'M} + \frac{\bar{\sigma}^2_\epsilon}{N'V'M} + \frac{\bar{\sigma}^2_\rho}{N'V'MR'}
\end{split}
\end{equation}

D-study variance allows us to estimate the standard error and 95\% CI: $\hat{\theta} \pm 1.96 \cdot \sqrt{\Var(\hat{\theta})}$, which accounts for all identified pipeline variance. The interval includes variance for generalizing to new judges and temperatures, which is why when the target is a fixed panel our approach is overly conservative (99 - 100\% coverage in SI Appendix, Section~\ref{si:underestimation_sim}, see also Section~\ref{si:underestimation_redraw}). 

Monte Carlo simulations (1,000 replicates) confirm that D-study intervention \emph{rankings} are highly reliable; under the default DGP, the correct top-1 intervention is identified in 100\% of simulations. Four assumption violations and a latent-ambiguity sweep each lead to $|\text{bias}| \leq 9\%$ (SI Appendix, Sections~\ref{si:dstudy_validation} and~\ref{si:latent_ambiguity}).

We use REML \citep{jiang1996reml} via \texttt{lmer} \citep{bates2015lme4} to estimate variance via a crossed random-effects specification. We consider both temperature and the model to be fixed, with every random facet and interaction as an independent term, for example, \texttt{(1|item) + (1|prompt\_variant) + (1|item:model)}. The value of variance of each of these terms corresponds to $\sigma^2$ in Equation~\ref{eq:tle}, which Equation~\ref{eq:dstudy_multi} divides by the number of levels a proposed design averages over. The full formula and pilot-design thresholds are in SI Appendix, Sections~\ref{si:framework} and~\ref{si:checklist}.

\paragraph{Choosing facets and judges.}
In practice, crossing through all possible factors rapidly becomes combinatorially expensive. We recommend running a small pilot study across the actual factors that vary in deployment and estimating the decomposition based on this to identify where the variance lies. In our applications, one or two factors usually accounted for most of the variance (most often judge choice), whereas temperature and further replications were much less important. Sampling parameters such as top-$p$ or top-$k$ could be included in the same fashion as temperature. On the other hand, switching between direct and chain-of-thought prompting, output format, or reasoning mode should be done with care since such modifications may change answer parsing or even construct of interest and thus cannot be used without passing the construct equivalence test discussed above, they are better treated as separate design factors. When selecting judges, we recommend starting with a small crossed pilot study using a broad set of plausible models, examining the judge marginals and leave-one-judge-out diagnostics (Section~\ref{sec:safety}), and comparing with human calibration labels if any. Since the recommendations depend on which prompts and judges are included in the pilot study, TEE is best applied as a diagnostic tool rather than a mechanical rule.

\section{Empirical Demonstrations}
\label{sec:empirical}

We apply the TEE decomposition across three empirical domains in the main text: binary safety classification (AILuminate), multiple-choice QA (MMLU), and pairwise model comparison validated against human votes (Chatbot Arena). For the safety and Arena demonstrations, we decompose variance across judges, treating each (prompt, SUT-response) pair as an item,\footnote{In Arena, each item is one battle, consisting of the user prompt and the two SUT responses being compared.} while in MMLU there is no judge, each benchmark question is an item, and we decompose variance across SUTs scored against the answer key (SI Appendix, Section~\ref{si:framework}). Section~\ref{sec:gaming} then quantifies the gaming surface single-configuration benchmarks expose. In the SI, we provide a political affinity annotation demonstration (Section~\ref{sec:propaganda}) with a 9-annotator human baseline, a Likert vs pairwise scoring comparison (Section~\ref{si:scoring_comparison}), and a pilot validation showing that a small sample recovers the qualitative D-study guidance of an 8$\times$ larger full run (Section~\ref{si:pilot}).

\subsection{Safety: Judge Disagreement Dominates}
\label{sec:safety}

We first illustrate the variance decomposition using AILuminate~v1.0 \citep{ghosh2025ailuminate}, a safety benchmark from MLCommons. The benchmark assesses whether LLM responses cross safety lines across 12 hazard categories (hate speech, self-harm, violent crime, etc.). Most items are unambiguously safe (94\% safe rate) and the variance concentrates on the 6\% of borderline cases where judges disagree about whether a refusal is adequate. We decompose the variance based on 36 demonstration prompts, using responses from Claude Opus 4.5, GPT-5.1, DeepSeek Chat~v3.1, and Grok 4.1 Fast (141 items after dropping 3 SUT refusals).%

\begin{figure}[htbp]
\centering
\begin{minipage}[t]{0.49\linewidth}
  \centering
  \includegraphics[width=\linewidth]{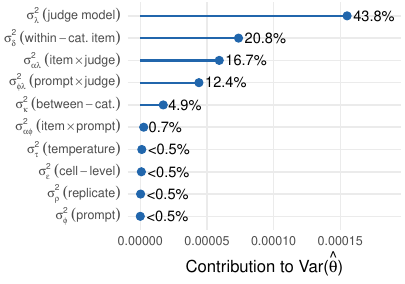}\\
  \footnotesize (a) Variance decomposition
\end{minipage}\hfill
\begin{minipage}[t]{0.49\linewidth}
  \centering
  \includegraphics[width=\linewidth]{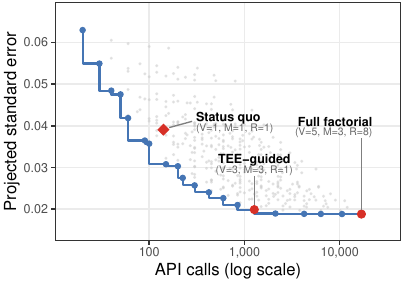}\\
  \footnotesize (b) Cost-efficiency frontier
\end{minipage}
\caption{Safety decomposition (a) and cost-efficiency frontier (b) for AILuminate. Judge-model variance dominates the decomposition in (a). A TEE-guided design that aggregates over diverse judges captures most of the available SE reduction at modest cost (b).}
\label{fig:safety_combined}
\end{figure}

Judge choice matters for safety rating variability and for item-level variance. The decomposition (Figure~\ref{fig:safety_combined}(a)) shows that the judge panel accounts for the largest share of variance (43.8\%), with the item$\times$judge interaction adding another 16.7\%. Both shrink as $1/M$ when the projection averages over $M$ judges. 

We also ran leave-one-judge-out estimates, which confirm that no one judge drives this pattern.  The three judge-related terms account for 73\% of the variance with the full panel, and 68 - 83\% after dropping any one judge (see also SI~\ref{si:lojo}). Of course, knowing \emph{which} judge is most accurate requires validation against gold standard labels (e.g., known-answer items, the hazard taxonomy, human labelled data). However, we can still get a lot of mileage from the decomposition alone --- a large judge mean offset implies we should recalibrate or remove a rater, while large item$\times$judge disagreement suggests closer review of specific cases.

Within-category item heterogeneity contributes 20.8\% and prompt$\times$judge 12.4\%, with each remaining component below 5\%. Replicate noise $\sigma^2_\rho$ and cell-level error $\sigma^2_\epsilon$ are large components at the item level, but enter $\Var(\hat\theta)$ divided by $N'V'H'M'R'$ and $N'V'H'M'$ respectively, so each contributes less than 0.5\% to the variance of the mean. Per-category variance spans more than an order of magnitude in replicate noise $\hat{\sigma}^2_\rho$, from $\hat{\sigma}^2_\rho = 0.001$ for specialized advice to $\hat{\sigma}^2_\rho = 0.033$ for sex crimes (SI Appendix, Section~\ref{si:pilot}).
Temperature and prompt main effects contribute negligibly. \footnote{Three-way interaction terms (item$\times$prompt$\times$temperature, item$\times$temperature$\times$judge) are statistically significant but substantively negligible, supporting the two-way interaction specification.}

The D-study then projects precision under any proposed design (any new combination of $N'V'H'M'R'$, etc.), enabling a cost-efficiency analysis. Figure~\ref{fig:safety_combined}(b) plots projected standard error against API cost for 360 candidate designs of the safety benchmark. The status quo design (141 items, one prompt, one judge, no replication) sits well above the efficient frontier; for the same budget, designs on the frontier that trade some items for judge diversity reduce the SE by 9\%. A TEE-guided design (141 items, 3 prompts, 3 judges, no replication) lowers the SE by 49\% at 9$\times$ the cost, whereas running the full factorial (5 prompts, 3 judges, 8 replications) yields only an extra 5\% reduction at 120$\times$ the cost. Beyond roughly 1,000 API calls, the curve flattens out with clear diminishing returns.

\subsection{MMLU: Item and Prompt Reallocation Roughly Halves Error}
\label{sec:mmlu}

Unlike our safety analysis, which centers on an LLM-as-judge/autorater layer, our MMLU benchmark analysis has no judge model at all. Because MMLU (Massive Multitask Language Understanding) \citep{hendrycks2021mmlu} consists of answers to multiple choice problems, we can measure estimation accuracy as well as coverage. We treat evaluation as a budget-allocation problem where the problem is how to allocate the API calls among number of items ($N$), prompt variants ($V$), and replications ($R$), given a fixed budget. We compare three approaches: (i) a \emph{naive} allocation ($V\!=\!1$, $R\!=\!3$) where each call is repeated three times, (ii) a \emph{standard} allocation ($V\!=\!1$, $R\!=\!1$) where the goal is to sample more items, and (iii) an allocation guided by TEE, which utilizes a D-study to choose between adding items, prompts, or replications at each budget level (see SI Appendix, Section~\ref{si:mmlu} for details).

\begin{figure}[htbp]
\centering
\begin{minipage}[t]{0.49\linewidth}
  \centering
  \includegraphics[width=\linewidth]{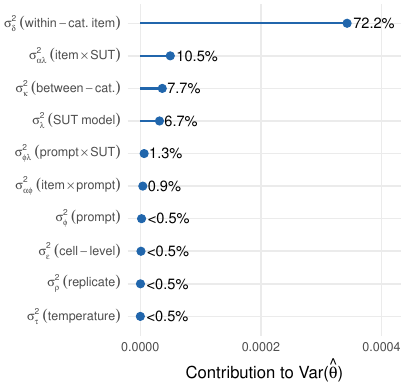}\\
  \footnotesize (a) G-study: variance decomposition
\end{minipage}\hfill
\begin{minipage}[t]{0.49\linewidth}
  \centering
  \includegraphics[width=\linewidth]{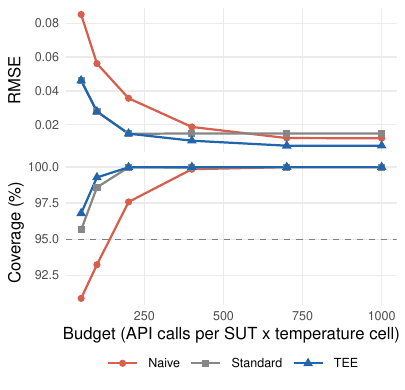}\\
  \footnotesize (b) D-study: budget allocation
\end{minipage}
\caption{TEE-guided estimation on MMLU roughly halves the standard RMSE at large budgets and maintains nominal 95\% CI coverage, while naive under-covers at the smallest budgets. (a)~Variance decomposition at the operational design ($N{=}200$, $V{=}4$, $M{=}3$ SUTs, $H{=}3$, $R{=}8$). (b)~RMSE against ground-truth accuracy (top) and 95\% CI coverage (bottom) by API call budget. 1,000 Monte Carlo replicates per SUT$\times$temperature cell.}
\label{fig:mmlu_combined}
\end{figure}

For the G-study (Figure~\ref{fig:mmlu_combined}(a)), items explain the majority of variance, with the contribution of heterogeneity within categories being 72.2\% of $\Var(\hat{\theta})$ and heterogeneity between categories contributing 7.7\%. Item$\times$SUT adds 10.5\% and SUT-model design sensitivity adds 6.7\%. In contrast, the three prompt variables combined explain only 2.7\%, with accuracy levels ranging from 83.9\% to 85.3\% across the four prompt variants.\footnote{A fifth prompt variant omitted the instruction to answer with a letter, so many responses were cut off. Because it fails the answer-format requirement in Section~\ref{sec:gstudy}, we exclude it here. Scoring these cut-off responses as wrong would increase the prompt-related share to 70\% of $\Var(\hat{\theta})$ (SI Appendix, Section~\ref{si:mmlu_parsing}).} Re-estimating the model with a logit link similarly finds item heterogeneity to be the largest component (71.4\%, with prompt terms at 6.1\%; SI Appendix, Section~\ref{si:glmm}).

Using these figures, the TEE D-study begins with sampling more items (80\% of variance) before moving on to sampling prompts (after using the 170 items left after a pilot of 30 items). Since the naive sampling approach uses two-thirds of its trials for replicating with hardly any information gain, it is 1.6 to 2.4 times more error-prone than TEE regardless of the budget size (Figure~\ref{fig:mmlu_combined}(b)). The standard strategy samples as much as items as it can; however, once all available items are sampled, its error remains unchanged. However, since TEE averages across prompt variants once the item pool is exhausted, it continues to lower its error until it achieves half of the error of the standard strategy's RMSE at larger budgets. It also yields well-calibrated confidence intervals, maintaining $\geq$97\% coverage throughout, whereas the naive CI omits prompt sensitivity and under-covers at the two smallest budgets (91\% and 93\%).

\subsection{Chatbot Arena: Coverage, Accuracy, D-Study Validation}
\label{sec:arena}

Chatbot Arena \citep{chiang2024chatbot} extends the domain to pairwise human choices. Gathering votes from people is slow and expensive, and (outside of Chatbot Arena) LLM judges are often used to score benchmarks. However, if we want to replace human raters with LLM judges, their scores shouldn't depend on prompt wording, judge models, etc. Thus we compare a TEE pipeline and a naïve single-judge approach with respect to 4,676 Arena matches (battles) selected from the publicly available \texttt{lmarena-ai/arena-human-preference-100k} dataset. We stratify by model and category (Creative Writing, Coding, Factual QA, and Persuasion). Each match has five prompt variants scored by three judges. We use both 1 - 5 Likert and pairwise scoring formats (see SI Section~\ref{si:arena} for details).

\begin{figure}[H]
\centering
\includegraphics[width=0.75\linewidth]{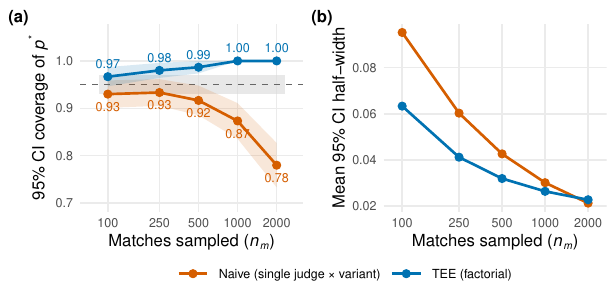}
\caption{Naive coverage on Chatbot Arena drops as the sample grows, while TEE maintains nominal coverage. (a)~Coverage versus match sample size $n_m$ for naive and TEE. (b)~Half-width of CI versus $n_m$. TEE approaches the appropriate floor ($\sigma_V^2/V + \sigma_J^2/J$) while naive keeps shrinking. Ground truth $p^* = 0.389$ is the mean across all 3 judges and 5 prompt variants in the Likert factorial, a finite-panel target. The TEE interval instead includes variance for generalizing to new prompts and judges, so it over-covers $p^*$ at large $n_m$. An interval for the finite-panel target would hold variants and judges fixed and bootstrap over matches.}
\label{fig:arena_coverage}
\end{figure}

A simulation experiment shows that TEE averaging recovers properly calibrated uncertainty estimates. Naive 95\% CIs based on a single (judge, prompt variant) cell have reasonable coverage at $n_m = 100$, but by $n_m = 2,000$ coverage drops to 78\% (Figure~\ref{fig:arena_coverage}).  The TEE CI remains at or above 95\%. Increasing the value of $n_m$ leads to a narrower interval centered on the average of one configuration, so more data actually means worse coverage.

TEE averaging also improves per-match agreement with the individual human vote (SI Appendix, Figure~\ref{fig:arena_accuracy}). Pooled across 4,676 matches, with both pipelines calling a winner on the same number of battles, TEE Likert raises AB-only agreement over single-judge Likert by $+5.5$ pp (72.5\% vs.\ 67.0\%) and TEE pairwise raises it over single-judge pairwise by $+1.3$ pp (69.6\% vs.\ 68.3\%, see SI Appendix Table~\ref{tab:arena_agreement}). %

A D-study fit on 80\% of matches projects per-category SE reductions of 0.53 - 0.59 from single-judge to TEE. Across the four categories, projected SE reductions track the observed Likert agreement lifts on all matches at Spearman $\rho = 0.80$ (n = 4; the D-study uses judge scores only; SI Appendix Figure~\ref{fig:arena_dstudy_vs_observed}).

Both findings matter when the goal is to score matches one at a time, as in training supervised models on Arena labels \citep{egami2024dsl} or filtering individual items.

\subsection{Single-Configuration Benchmarks Vulnerable to Benchmark Hacking}
\label{sec:gaming}

\citet{singh2025leaderboard} document 27 private Llama-4 variants tested on Chatbot Arena before launch, with only the top scorer published (\emph{benchmark hacking}), which constitutes best-of-$K$ exploitation of pipeline noise. Some ways it can happen are that, for leaderboards, many models are submitted to the leaderboard using different names, and the top performing one is reported. In open-source benchmarks, they can conduct the test inside using various forms of prompts and scoring configurations until they get a favorable score. The expected inflation is $\E[\max(Z_1,\ldots,Z_K)] \cdot \sigma_{\text{pipeline}}$, where $\sigma_{\text{pipeline}}$ is the standard deviation of the pipeline noise decomposed above. Prompt formatting alone can produce up to 76 accuracy points of spread on identical models \citep{sclar2024sensitivity}, suggesting the gameable surface is broad in practice.

\begin{figure}[htbp]
\centering
\includegraphics[width=0.8\linewidth]{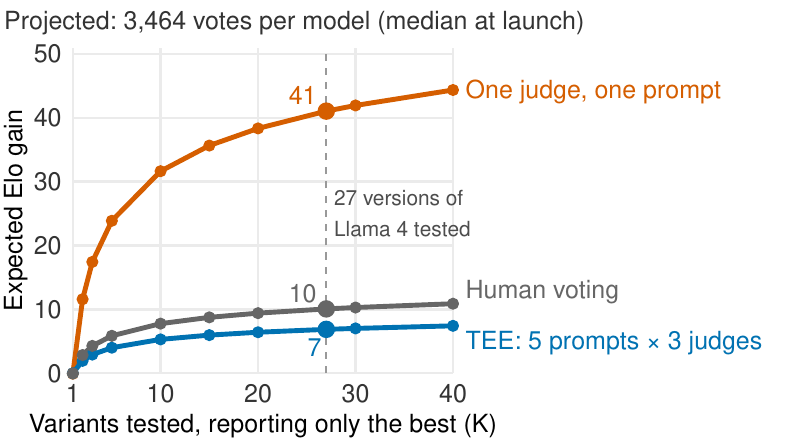}
\caption{Best-of-$K$ gaming surface on Chatbot Arena across pipeline configurations. ``Gaming surface'' is the expected Elo boost a developer earns by submitting $K$ model variants and reporting only the highest-scoring one, $\E[\max(Z_1,\ldots,Z_K)] \cdot \sigma_{\text{pipeline}}$. The dashed line marks $K{=}27$, the number of private Llama-4 variants tested on Chatbot Arena \citep{singh2025leaderboard}. Projected to 3,464 votes per model, the median at launch on the live leaderboard, from bootstrapping 4,676 Arena matches (SI Appendix, Section~\ref{si:gaming_projection}); on the 4,676 matches themselves, the values at $K{=}27$ are 56, 45, and 32 Elo.}
\label{fig:gaming_surface}
\end{figure}

We quantify the gaming surface in the context of Chatbot Arena by bootstrapping 4,676 matches across pipeline configurations. On these 4,676 matches (a median of 163 per model), human voting alone already exposes a 45-Elo surface at $K{=}27$, large enough to flip rankings between adjacent models. A single-configuration LLM-judge replacement ($V{=}1$, $M{=}1$) increases the surface to 56 Elo, since a developer can target both the LLM-judge configuration and the underlying voter sample. Averaging across either prompts or judges alone recovers the human baseline ($V{=}5$, $M{=}1$: 46 Elo; $V{=}1$, $M{=}3$: 46 Elo). TEE-LLM ($V{=}5$, $M{=}3$ averaging) cuts the surface to 32 Elo, about 30\% below human voting on the same matches. Voter and match sampling shrink as a model collects votes, but a single configuration's judge and prompt noise does not: at 3,464 votes per model, the median at launch on the live leaderboard, we project a surface of 10 Elo for human voting, 41 for a single configuration, and 7 for TEE-LLM (Figure~\ref{fig:gaming_surface}; SI Appendix, Section~\ref{si:gaming_projection}). Judge main effects account for the largest share of gameable LLM variance (Table~\ref{tab:gaming}), the component a developer most directly controls when picking judge models.

The best-of-$K$ simulation comprises just one facet of the manipulation surface. \citet{singh2025leaderboard}'s analysis finds that providers can retract scores from unfavorable runs, that proprietary providers received a disproportionate share of data (61.4\% of all battle data), and that access to Arena-specific data boosted performance by up to 112\% on the Arena distribution without corresponding gains on external benchmarks.

\section{Discussion}
\label{sec:discussion}

Across the empirical demonstrations, naive error estimates substantially understate measurement uncertainty. Naive CIs omit judge disagreement, prompt sensitivity, and their interactions, which can lead evaluators to mistake pipeline noise for genuine benchmark capability differences or generalizable patterns in LLM-annotated research. The gap between naive and TEE-informed error estimates corresponds to the relative variance between pipelines, which scales with both subjectivity and complexity. The data we analyze from Chatbot Arena, MMLU, and our propaganda audit (SI Appendix) further show that failure to aggregate over these noise sources also reduces accuracy. %

\begin{figure}[htbp]
\centering
\includegraphics[width=\linewidth]{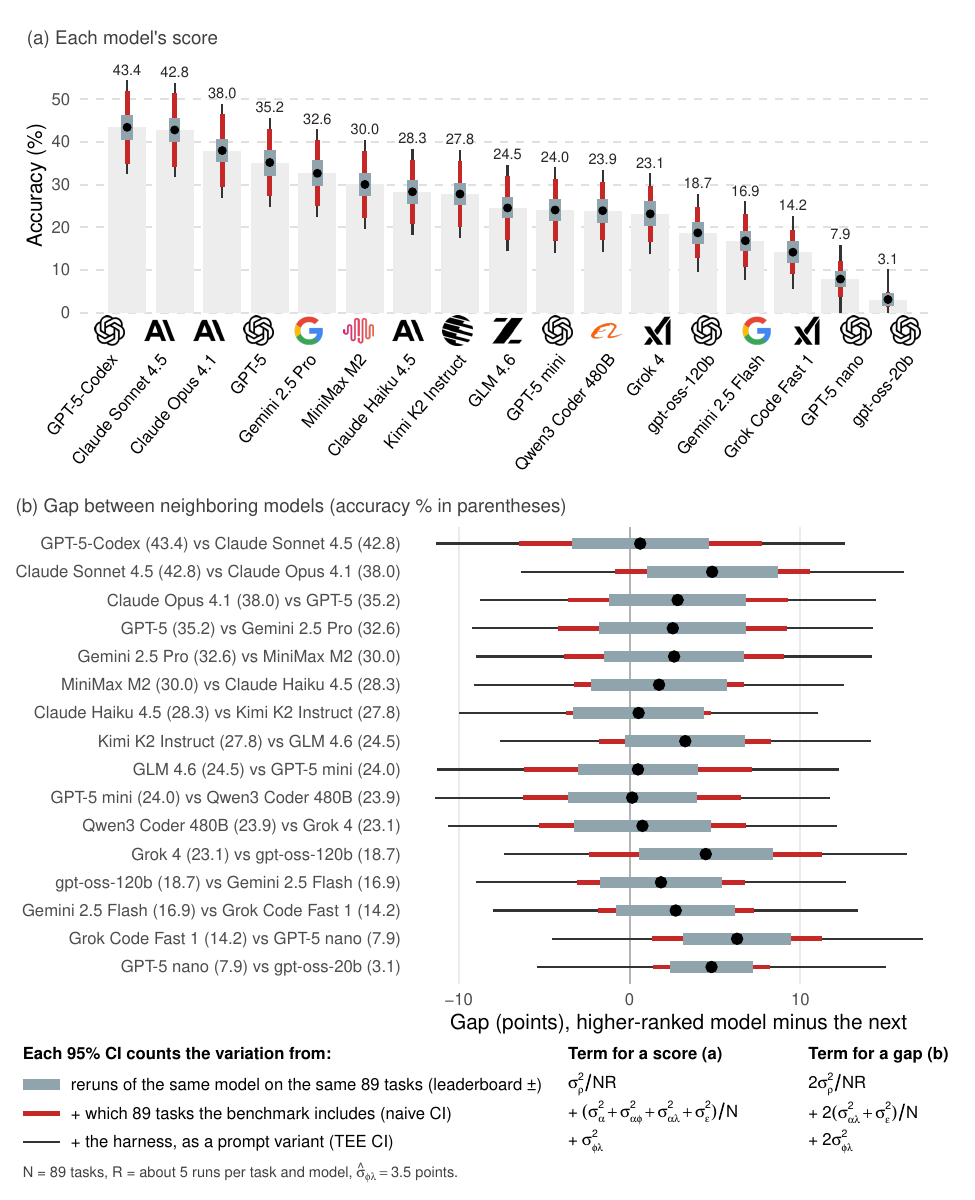}
\caption{Once we account for harness variance (prompt variance under TEE), all neighboring models on the Terminal-Bench 2.0 leaderboard have substantial CI overlap. (a)~Each model's score under the Terminus 2 harness (bar and black dot). (b)~The gap between each model's score and the next (black dot; accuracies in parentheses). Both panels show three nested 95\% CIs. The thick grey bar is the leaderboard's interval, which counts only reruns of the same 89 tasks. The red line is the naive interval over tasks, paired in (b). The thin black line is the TEE interval, which treats harnesses as prompt variants and adds the model$\times$harness variance $\sigma^2_{\phi\lambda}$, estimated from settings-matched cells ($\hat\sigma_{\phi\lambda} = 3.5$ points). Data: 22,198 trials of 17 models under six harnesses \citep{merrill2026terminalbench}; details in SI Appendix, Section~\ref{si:tbench}.}
\label{fig:tbench}
\end{figure}

In fact, some benchmarks report intervals that are even more anticonservative than the naive interval. SWE-bench \citep{jimenez2024swebench} and Tau-bench \citep{yao2024taubench} report scores with no interval at all. Other benchmarks report confidence intervals that are narrower than the naive interval, because they hold the evaluation items fixed and only account for variation between repeated runs (replicates). And, in its system cards, OpenAI resamples ``model attempts per problem'' and notes that this ``can lead to overly tight confidence intervals'' \citep{openai2025gpt5card}. MLE-bench and PaperBench report the standard error across replicates \citep{chan2024mlebench, starace2025paperbench}, and Vals.ai similarly estimates replicate error \citep{vals2026methodology}. Terminal-Bench 2.0 estimates error based on the spread of trials within each task, on their leaderboard and in their paper \citep{merrill2026terminalbench}, which again assumes fixed items. In our reanalysis of Terminal-Bench 2.0, the published intervals separate 4 of the neighboring model pairs, while the TEE intervals overlap on all 16 pairs (Figure~\ref{fig:tbench}; SI Appendix, Section~\ref{si:tbench}). Readers should treat rerun-only intervals as lower bounds on the uncertainty. 

Pipeline variance can also serve as a gaming surface in benchmarking and in research. We show that submitting best-of $K{=}27$ variants (as documented in \citet{singh2025leaderboard}) inflates Chatbot Arena Elo scores by 56 points, while averaging over multiple judges and prompts cuts that surface by 43\% (\S\ref{sec:gaming}). The same vulnerability extends to LLM-annotated research, where pipeline variance maps onto labeling error and predicts misclassification against the 9-rater human consensus in our propaganda demonstration ($r = -0.68$, SI Appendix, Section~\ref{sec:propaganda}), consistent with the configuration- and adversarial-search reversals \citet{baumann2025llmhacking} document across 37 LLM-annotation tasks. The implications extend beyond model capability overestimates, as benchmarks like SWE-Bench, LegalBench, and HealthBench inform estimates of labor substitution and productivity gains \citep{eloundou2023gpts} associated with widespread LLM adoption.

How should researchers and benchmark designers respond in practice? Across our applications, whenever a pipeline includes an LLM judge layer, our D-study projections indicate that averaging across multiple judges reduces variance more than any other single change, followed by adding items on AILuminate and adding prompt variants on Arena. Following those projections improves both CI coverage and agreement with human ground truth. More broadly, because almost all evaluation claims aim to generalize across a domain of items, sampling more items is consistently valuable, and results should not rest on a single judge model or prompt template. Even though the variance percentages provided by the rankings here are helpful guidelines, their actual values vary depending on the application, and for this reason, doing a pilot study in the target pipeline will prove extremely useful.

Because benchmark results and LLM annotations are used to carry out safety assessments and empirical studies and make projections about the economy or labor market, underestimating the variability of the pipeline will lead to distorted conclusions further down the line. Conducting a pilot study in the TEE framework will help achieve that at much less expense than in a full factorial design.

\section{Limitations}
\label{sec:limitations}

TEE has a number of simplifying assumptions worth noting. For example, when a model emits reasoning traces, the variance they generate is not attributed under the current framework and instead inflates $\sigma^2_\epsilon$. What's more, when the scoring function relies on an LLM-as-judge, its variance may compound with the SUT's \citep{haldar2025roulette, chehbouni2025neither}, and SUT$\times$judge interactions aren't currently captured in our decomposition (SI Appendix, Section~\ref{si:coverage}). 

Perhaps more importantly, TEE cannot say whether the items (eval prompts) truly cover and represent the constructs of interest, which of course limits generalizability. A toxicity benchmark that omits coded hate speech will generate precise and incomplete estimates. Total survey error has long called this coverage error \citep{sen2019tedon}.

Last, our demonstrations comprise English-only tasks with API-served models, and do not test ambiguous tasks that require open-ended reasoning.

\section{Conclusion}
\label{sec:conclusion}

If they are provided at all, variance estimates in LLM-as-judge or autorater evaluations are anti-conservative. This means our frontier model evaluations overstate precision, while naively scored benchmarks let developers exploit measurement error in a way analogous to p-hacking or overfitting. Our TEE framework not only helps detect this issue but lets researchers estimate variance components from a small pilot study, estimate confidence intervals with valid coverage, and predict how design modifications will help in reducing error (SI Appendix, Section~\ref{si:pilot}). As LLM evaluations become more important for practical applications (model deployment, safety criteria), research, and estimates of the impact of AI on the economy and the labor market, it is vital to take pipeline uncertainty into account.

\begin{ack}
Thanks to Chris Barrie, Ben Guinaudeau, Aaron Kaufman, Brandon Stewart, Kylan Rutherford, and participants at the 2026 CSMAP conference for helpful feedback. Thanks to Peter Mattson, Bennett Hillenbrand, Andrew Gruen, Rebecca Weiss, Hannah Waight, Molly Roberts, Brandon Stewart, Eddie Yang, Ben Guinaudeau, Melina Much, Chris Barrie, Joe Higton, Kylan Rutherford, Jennifer Allen, Jonathan Nagler, and Josh Tucker for helpful early conversations about the problem of confidence intervals in LLM evals, benchmarks, and LLM-as-judge measurement problems, which helped inspire this paper.

MLCommons funded the author's salary for this work. Thanks also to NYU's Center for Social Media, AI, and Politics for funding this research.

The author has personal financial interests in AI-related companies, in particular Nvidia, Alphabet, Microsoft, and Taiwan Semiconductor. The author worked at Facebook (now Meta) in various capacities from 2011 to 2015 and 2018 to 2020 and at Twitter (now X) from 2021 to 2023, and contracts for MLCommons, a 501(c)(6) non-profit that releases AI benchmarks, including AILuminate, which this paper analyzes (2026 to present). The author has accepted a position at Google DeepMind. For an unrelated project, the author received a 2010 Google Research Award for research on ``Social cues and reliability in content selection and evaluation.''
\end{ack}

\bibliographystyle{plainnat}
\bibliography{references}

\end{document}